\documentclass[prodmode,acmjdiq]{acmsmall} 

\usepackage[ruled]{algorithm2e}

\SetAlFnt{\small}
\SetAlCapFnt{\small}
\SetAlCapNameFnt{\small}
\SetAlCapHSkip{0pt}
\IncMargin{-\parindent}

\acmVolume{0}
\acmNumber{0}
\acmArticle{0}
\acmYear{2016}
\acmMonth{2}

\setcopyright{rightsretained}

\doi{0000001.0000001}

\issn{1234-56789}

\usepackage{amsmath}
\usepackage{etoolbox}
\usepackage{bm}
\usepackage{mathtools}
\usepackage{subfigure}
\usepackage{xspace}
\usepackage{color}

\newcommand{\MyMapTemplatePrefixc}[4]{\expandafter#1\csname#3#4\endcsname{#2{#4}}} 
\forcsvlist{\MyMapTemplatePrefixc {\def} {\mathcal}{c}} {A,B,C,D,E,F,G,H,I,J,K,L,M,N,O,P,Q,R,S,T,U,V,W,X,Y,Z}  

\newcommand{\MyMapTemplatePrefixtb}[5]{\expandafter#1\csname#4#5\endcsname{#2{#3{#5}}}} 

\newcommand{\MyMapTemplateNoPrefix}[3]{\expandafter#1\csname#3\endcsname{#2{#3}}}
\forcsvlist{\MyMapTemplateNoPrefix {\def} {\mathbf} } {0,1,a,b,c,d,e, f, g, h, i, j, k, l, m, n, o, p, q, r, s, t, u, v, w, x, y, z} 
\forcsvlist{\MyMapTemplateNoPrefix {\def} {\mathbf} } {A,B,C,D,E,F,G,H,I,J,K,L,M,N,O,P,Q,R,S,T,U,V,W,X,Y,Z}  

\newcommand{\zheng}[1]{}

\newcommand{\SAVE}[1]{}

\begin{document}

\markboth{Z. Xu et al.}{Exploiting Lists of Names for Named Entity Identification}

\title{Exploiting Lists of Names for Named Entity Identification of Financial Institutions from Unstructured Documents}
\author{ZHENG XU
\affil{University of Maryland}
DOUGLAS BURDICK
\affil{IBM Research}
LOUIQA RASCHID
\affil{University of Maryland}
}

\begin{abstract}
There is a wealth of information about financial systems that is embedded in
document collections.
In this paper, we focus on a specialized text extraction task for this domain.
The objective is to extract mentions of names of financial institutions, or
FI names, from financial prospectus documents,
and to identify the corresponding real world entities,
e.g., by matching against a corpus of such entities. The tasks are Named Entity Recognition
(NER) and Entity Resolution (ER); both are well studied in the literature.
Our contribution is to develop a rule-based approach that will exploit
lists of FI names for both tasks;
our solution is labeled Dict-based NER and Rank-based ER.
Since the FI names are typically represented by a root, and a suffix that modifies the root,
we use these lists of FI names to create specialized root and suffix dictionaries.
To evaluate the effectiveness of our specialized solution for extracting FI names,
we compare Dict-based NER with a general purpose rule-based NER solution, ORG NER.
Our evaluation highlights the benefits and limitations of specialized
versus general purpose approaches, and presents additional suggestions for
tuning and customization for FI name extraction.
To our knowledge, our proposed solutions, Dict-based NER and Rank-based ER,
and the root and suffix dictionaries,
are the first attempt to exploit specialized knowledge, i.e., lists of FI names,
for rule-based NER and ER.
\end{abstract}

%
%


\terms{Design, Algorithms, Performance}



\acmformat{Zheng Xu, Douglas Burdick,
and Louiqa Raschid, 2016.
Exploiting Lists of Names for Named Entity Identification of Financial Institutions from Unstructured Documents.
}


\begin{bottomstuff}
This work is partially supported by the National Science Foundation
grant DBI1147144 and National Institute of Standards grant 70NANB15H194.

Author's addresses: Z. Xu {and} L. Raschid, University of Maryland; D. Burdick,
IBM Research.
\end{bottomstuff}

\maketitle

\section{Introduction}

The behavior of financial contracts and systems can be better modeled and understood when
there is improved transparency and detailed knowledge of the underlying
complex financial supply chains.
An example is the behavior of the system comprising
US residential mortgage backed securities, resMBS.
This system combined with the subprime mortgage crisis to lead to the 2008
U.S. financial crisis.
The rich financial network that describes this supply chain, i.e.,
the financial institutions and the role(s) that they play on resMBS contracts, is deeply
embedded in prospecti that usually consist of
hundreds of pages of semi-structured text.
While these prospecti are public and filed with the Securities and Exchange Commission (SEC),
there has been limited activity to harvest them to create financial datasets.
Some proprietary datasets that describe the resMBS supply chain are available for a fee
from vendors; they focus on the performance of individual prospecti and not on the supply chain.

The absence of such datasets prevents the types of financial big data analytics
that will be very useful to both regulators and investors who have an interest in
real estate and mortgage capital markets ~\cite{burdick2014dsmm,burdick2016dsmm,xu2016dsmm}.
This gap was made evident during, and in the aftermath of, the 2008 crisis
when regulators and analysts had to make decisions in the absense of knowledge
about systemic risk across this supply chain.
The information extraction and data management tasks that are required to create
financial big data collections such as the resMBS dataset
present an interesting challenge to data scientists.


Information extraction (IE) refers to the problem of extracting structured information
from unstructured text. It is a vital part of creating big data collections.
Methods for IE have gained significant traction in natural language processing,
information retrieval and database and data analytics research ~\cite{chiticariu2010sigmod}.
Within IE, recognizing information units like names of persons or places or organizations
is known as named-entity recognition (NER) ~\cite{nadeau2007ner}. Matching and resolving
these mentions of named entities against a database of concepts is known as,
among other alternate labels, entity resolution (ER)~\cite{getoor2012vldb}.

Methods for NER can be classified into the following three categories: rule-based;
machine learning-based; hybrid~ \cite{chiticariu2013emnlp}.
Statistical machine learning approaches are widely used in the academic community.
However, recent rule-based approaches ~\cite{chiticariu2010emnlp} developed on
top of the System T declarative platform ~\cite{chiticariu2010acl}
achieved state-of-the-art accuracy on the NER task.
In comparison to machine learning approaches ~\cite{florian2003naacl,minkov2005emnlp},
the rule-based approach only requires moderate efforts for manual customization
of rules and minimal labeled data. It also benefits from the ability to provide
a better explanation of successes and errors.

Rule-based NER has been applied to financial documents ~\cite{burdick2011deb,hernandez2010unleashing}.
Those efforts relied on a general purpose NER for organizations, ORG NER, which
will be described later.
When applying information extraction for a specific application
domain, customization is a standard but nontrivial modification to improve performance.
Machine learning-based approaches may require additional labeled data and
a retraining of the model~\cite{ritter2011emnlp}.
Rule-based approaches may require a manual redesign of the rules ~\cite{chiticariu2010emnlp}.

In this paper, we propose a specialized rule-based solution with a focus on
the extraction of mentions of the names of financial institutions, i.e., the extraction
of FI names.
Our innovation is to exploit lists of FI names,
and to customize a two-part solution, Dict-based NER and Rank-based ER.
Dict-based NER and Rank-based ER are built upon a general purpose algebraic information
extraction system,
System T, and its programming language AQL ~\cite{chiticariu2010acl}.
The benefits of using System T include the rule-based paradigm and the
scalability of using a distributed system.

We combine multiple name lists from several sources for Dict-based NER.
In contrast, we utilize a smaller targeted list of names for Rank-based ER.
\footnote{We used several noisy lists, e.g., from the SEC. A more
targeted list was obtained from ABSNet,
\url{www.absnet.net/},
a vendor providing data and
analytics for a range of asset backed securities.}
We observe that FI names can typically be split into a root fragment and a suffix.
The root, e.g., "Wells Fargo", can distinguish among financial institutions.
The suffix typically identifies the type of institution and are usually common among a lot of FIs, for example, "Bank", "N.A."
or "National Association".
A root dictionary and a suffix dictionary are explicitly generated from the lists of FI names,
and Dict-based NER will utilize a dictionary matching function to perform extraction based on the dictionaries.
For Rank-based ER, we develop a scoring function to select the best
matches against a corpus of FI entity names. The concept of distinguishing root and suffix of the FI names are essential in both modules, as will be discussed.

We evaluate the effectiveness of Dict-based NER and Rank-based ER by extracting
names of financial institutions from a collection of over 5000 resMBS prospecti
that were filed with the SEC between 2000 and 2008
\footnote{We downloaded the documents from the SEC website \url{http://www.sec.gov/}}.
Dict-based NER recognizes and extracts the mentions of FI names and
Rank-based ER links these extracted mentions against a corpus of FI entity names from ABSNet.
We used the general purpose ORG NER as a control for comparison.
The evaluation was manually validated over a sampled subset of prospecti.

\zheng{I do not like the following paragraph, we should move it to a related work section, if we will have one. You can decide since I do not want to justify the comparison between Dict NER and ORG NER.}
After appropriate tuning, the general purpose ORG NER yielded good precision and recall.
We observed that most errors for ORG NER appeared to be incomplete extractions.
Figure \ref{fig:intro} shows some fragments from financial prospecti.
The reasons for the errors made by ORG NER and the challenge of tuning ORG NER
for the specialized task of FI name extraction are discussed in a later section.
The specialized Dict-based NER improved on the performance of ORG NER.
However, it was limited in its ability to generalize the approach
beyond the entries provided in the dictionaries.
This limitation was particularly noted when Dict-based NER encountered
a prospectus from an FI, where the training prospecti did not
include examples from that FI, i.e., a previously unseen FI.
In this case, the root and suffix dictionaries may not have entries that could
help in the matching task.  The details are presented in the paper.

\zheng{Revised. I want a larger community to be attracted and interested. I want to claim using extra sources as a contribution. And make NER/ER people look at this paper. It would be better if we can find some NER/ER people to look at the paper and justify our claims.}
We expect that our approach to be widely applicable across many types of
financial documents. Moreover, our practical approach demonstrates the benefits of exploring and exploiting extra sources, such as lists of names, for domain specific tasks in information extraction.
The idea of splitting a name of an entity into distinguishable part and common part can also be utilized across other application domains.
The proposed approach is intuitive and unsupervised, which makes it extremely easy for users to get familiar with. 
It falls within the scope of rule-based approach, which needs no labeled training data. Comparing with rule-based general purpose approaches, the required manual efforts for customization are relatively little.  

Two key conclusions are that exploiting lists of FI names,
and splitting functional dictionaries (explicitly or implicitly), a la the Dict-based NER and Rank-based ER method,
can improve on a general purpose solution.  However, these specialized solutions
have limitations with respect to generalizing their capability.
This is particularly the case when handling a prospectus from an unseen FI.
A comprehensive solution may
require both specialized and general purpose solutions. There is also a need for
additional extensions, e.g., a regular expression based customization.

This paper is organized as follows:
Section \ref{sec:solution} provides a motivating example and
provides an overview of the proposed specialized solution for the extraction of
financial institutions, which includes Dict-based NER  and Rank-based ER. 
The details of Dict-based NER  and Rank-based ER are described in section \ref{sec:ner} and section \ref{sec:er}), respectively.
Section \ref{sec:exp} presents the results of an extensive evaluation and
manual validation; Dict-based ER is compared against the general purpose ORG NER.

\zheng{
\section{related work}
todo
}

\section{Proposed Solution}
\label{sec:solution}

In this section, we describe our approach for NER and ER,
to extract and resolve FI names from unstructured resMBS prospecti.
We use some examples from Figure \ref{fig:intro} to illustrate the challenges.
As mentioned earlier, our innovation is to exploit lists of FI names
as an external resource, for both NER and ER tasks.
Based on the observation that the names of financial institutions can typically
be split into a root fragment and a suffix,
we exploit a root dictionary and a suffix dictionary for Dict-based NER and Rank-based ER.

\subsection{Motivating Example}

\begin{figure}[tbhp]
\centerline{\includegraphics[width=\linewidth]{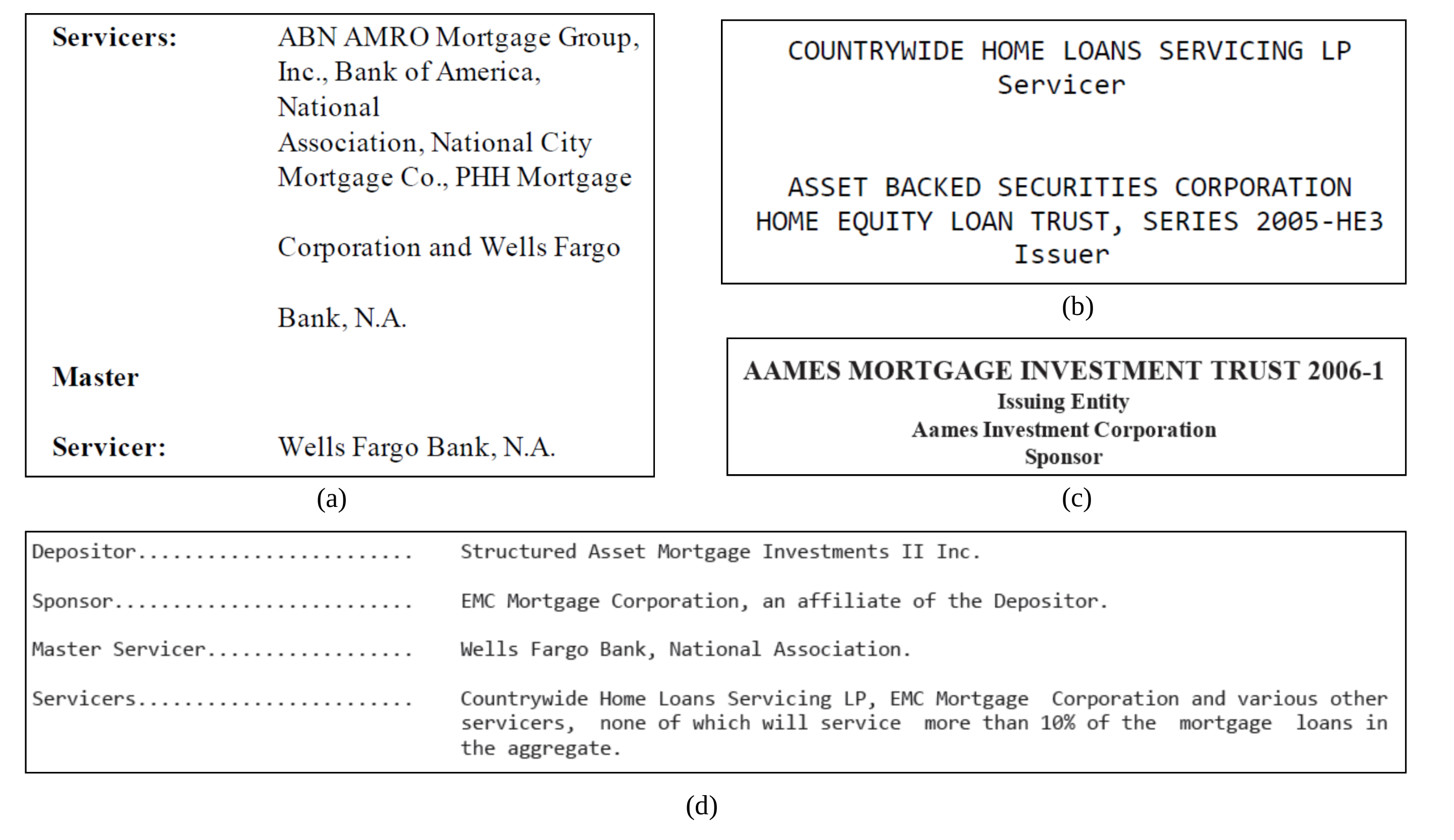}}
\caption{Text fragments from financial contracts illustrate the challenges of NER.
(a) The text lacks NLP features and names are broken over several lines.
(b) and (c) illustrates the capitalization of names of financial institutions and
the inclusion of numerical values in the name.
(a) and (d) illustrate different suffix terms for the same financial institution,
e.g., Wells Fargo Bank, N.A. and Wells Fargo Bank, National Association.}
\label{fig:intro}
\end{figure}

Figure \ref{fig:intro} shows some examples of text from prospecti.
The task of extraction of the mentions of FI names is challenging for the following reasons:
\begin{itemize}
\item
Financial institutions may have long complex names and most general purpose
NER approaches may fail to handle such names.
For example the issuer (issuing entity) of a resMBS contract is often
a trust that is formed for this purpose. Its name may include a numeric suffix
that is not typically expected in a name.
This is illustrated in Figure ~\ref{fig:intro} (b) and (c).
\item
The complex layout of the resMBS prospectus, which is a legal document,
makes it difficult to identify mentions of the financial institutions.
There are several templates defining the structure of the resMBS prospectus and
they often do not provide obvious tags that can be used for the NER task.
\item
The financial institution name often appears in an individual line that may
be free of additional text so that it lacks context, natural language features and structure tags.
Further, due to the abnormal format of some prospecti, names can break across
several lines. Names are also sometimes capitalized.  
\footnote{Financial contracts use capital letters in paragraphs to emphasize the words.} 
Those specific formatting issues are difficult for conventional NER.
\item
Similarly, entity resolution (ER) is difficult since mentions for the same institution
may vary widely.
A financial institution may be mentioned using different names and/or abbreviations,
e.g., "Wells Fargo", "Wells Fargo Bank", "Wells Fargo Bank, N.A." and
"Wells Fargo Bank National Association".
This is illustrated in Figure ~\ref{fig:intro} (a) and (d).
In this case, all of these names may represent FIs that are affiliated with
a single parent or focal FI.
\end{itemize}

\subsection{System Overview}

\begin{figure}[tbhp]
\centerline{\includegraphics[width=\linewidth]{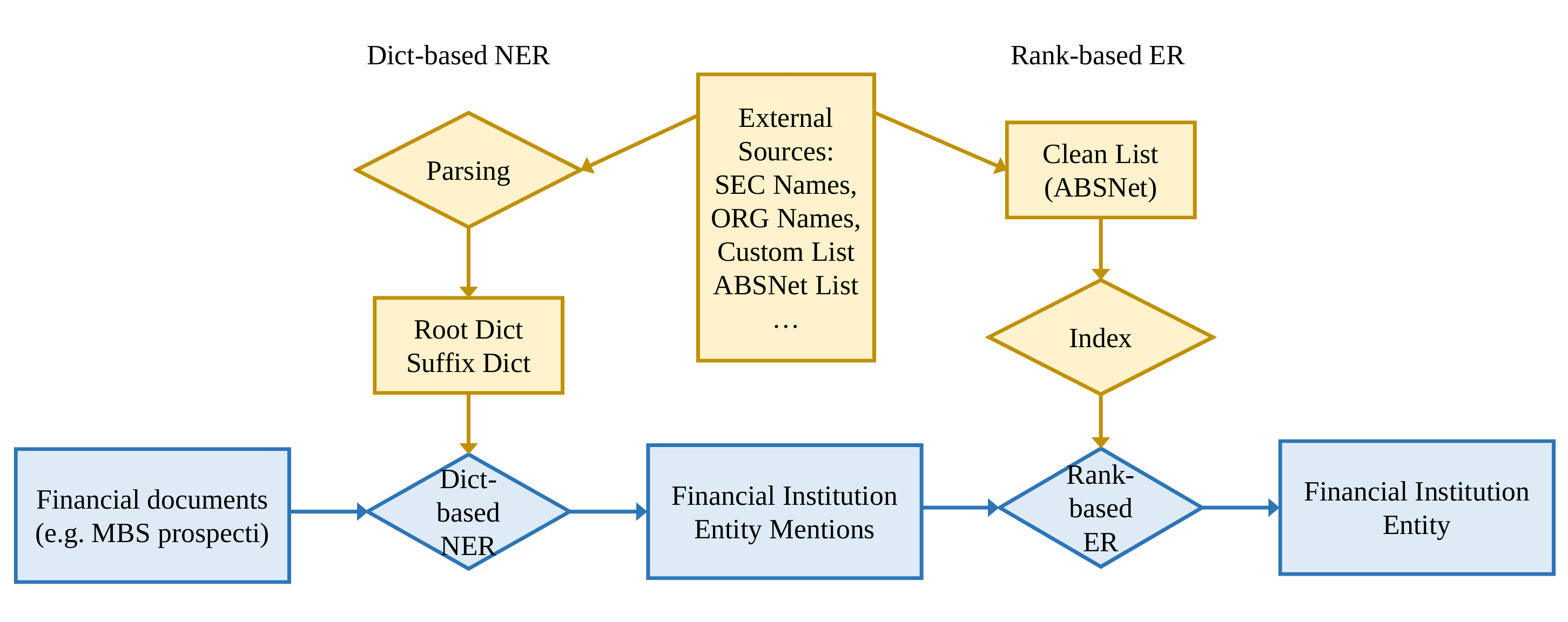}}
\caption{System overview of a specialized financial institution (FI) extractor
that comprises
Dict-based NER and Rank-based ER. Dict-based NER will extract mentions of
FIs and Rank-based ER will resolve these mentions against a targeted list of FIs.}
\label{fig:overview}
\end{figure}

We summarize the pipeline of Figure ~\ref{fig:overview}.
The NER task will extract mentions of FIs from input resMBS prospecti.
ER will resolve those mentions against a targeted list of FIs.
Dict-based NER is based on dictionary matching.
While matching, we first use the root dictionary to extract the distinct
root fragment of the name of the financial institution.
We then append the suffix to the root to generate the complete name of
the financial institution.
The two dictionaries are generated from external sources of name lists
which may be noisy and incomplete.
We carefully design the parsing and dictionary matching task to be tolerant of the
noisy name lists, and to improve recall, i.e., the coverage of names of FIs.
We note that the robustness and scalability of dictionary matching is due to
the benefits that come from using the System T platform ~\cite{chiticariu2010acl}.

By using separate root and suffix dictionaries, and the ability to combine
knowledge from both, we extend the capability of our approach.
We can use a combination of known root and suffix values to
infer new names of FIs, to handle abbreviations, etc.
For example, by combining root and suffix values from "Wells Fargo Bank" and "Countrywide MBS",
we could also infer and extract additional
names of FIs including "Wells Fargo MBS" and "Countrywide Bank".
We discuss the details of Dict-based NER in Section ~\ref{sec:ner}.

The goal of the ER task is to resolve mentions and map (one or more) mentions
to a single financial institution (FI).
For example, "Wells Fargo Bank", "Wells Fargo Bank, N.A.", and "Wells Fargo Bank National Association"
should all (potentially) be mapped to "Wells Fargo".
Rank-based ER exploits a corpus of a targeted and normalized list of names of FIs.
We consider each FI name in this list to be a document, and we use a bag-
of-words model on the corpus of FI names for this task.

We develop a scoring function that is inspired by term frequency and inverse
document frequency (TF-IDF).
Rank-based ER also uses several heuristics based on the observed properties of FI names.
For each FI mention, the scoring function will be used to create a ranking and to find the best match from the corpus.
We use a threshold on the score to retain valid matches.
Rank-based ER uses an inverted index and is efficient and easy to parallelize.
Unlike NER, the root and suffix fragments of the name are not separated and are
incorporated into the scoring function.
We discuss the details of Rank-based ER in Section ~\ref{sec:er}.

\section{Dict-based NER}
\label{sec:ner}
In this section, we present the details of the Dict-based NER module.
This includes the tasks of dictionary generation and dictionary based matching.
Dict-based NER makes the following assumptions:
\begin{itemize}
\item
The FI names are composed of a distinguishable part, i.e. root fragment and a modifier, i.e., suffix.
The root fragment tends to be distinct. The suffix does not show much
variation across multiple mentions of the FI.
\item
An (almost) complete list(s) of formal names for financial institutions (FIs)
is available so that we can effectively construct dictionaries from the list(s).
\item
A relatively similar version of the formal name of an FI will appear at least
once in the document, so that Dict-based NER can use the dictionary to extract
at least one mention of the FI that will match the formal name of the FI.
\end{itemize}

\subsection{Dictionary Generation}
\label{sec:dict_gen}


Next, we present details on dictionary generation from a lists of names.
We use lists from the following three sources:
\begin{itemize}
\item
A list of organization names from the SEC; it included 174851 unique names.
This list was noisy since it contained the names of many organizations that are not FIs.
Of greater concern is that it was incomplete.
\item
A list that was utilized by ORG NER on SystemT ~\cite{chiticariu2010emnlp};
it contained 6874 names.
\item
A small customized list of FI names that was manually constructed using fifteen prospecti
\footnote{We use 15 example files from the following 15 distinct FIs:
BANC\_OF\_AMERICA, BEAR\_STERNS, COUNTRYWIDE, CREDIT\_SUISSE\_FIRST\_BOSTON,
FIRST\_HORIZON, GOLDMAN\_SACHS, INDYMAC, MASTR, MERRILL\_LYNCH, MORGAN\_STANLEY,
RALI, RESIDENTIAL\_FUNDING, SASC, WAMU, WELLS\_FARGO}.
This list of FI names included approximately 50 names and was very
valuable to improve precision.  Unfortunately this list was also incomplete.
\end{itemize}

Using the fifteen prospecti as a guide, we developed the following heuristics to
generate entries in the root and suffix dictionaries:

\begin{itemize}
\item
Remove short text that occurs after '\textbackslash', '/', '\#',
e.g.,  B HANAUER \& CO /BD'.
This rule helps overcome several instances of noisy names, in particular when using the SEC list.
\item
Skip names that have less than 5 characters, such as 'O'.
\item When a name includes a comma, in the following format: 'A, B, C, ..., D, E',
we add the tokens 'A', 'A, B', 'A, B, C' and, 'A, B, C\ldots, D' to the root dictionary.
Further, we may add the tokens 'B', 'C', \ldots, 'D' to the root dictionary if the
token is long.
\item
In this case, we also add ',E', ',D', \ldots, ',C', and ',B' to the suffix dictionary.\\
For example, from 'SOUTHEAST INVESTMENTS, N.C., INC.', we add
'SOUTHEAST INVESTMENTS' and 'SOUTHEAST INVESTMENTS, N.C.' to the root dictionary.
We further add ', N.C.' and ',INC.' to the suffix dictionary.
\item When a name does not include a comma, add the last token of the name to the
suffix dictionary, if the name does not contain the token 'OF'.
Add the last two tokens to the suffix dictionary if the last token is short or contains
a lot of digits.
\item
Add the whole name, the name without the suffix, the name without the first token,
etc. to the root dictionary; this will help to improve recall.
For example, from 'J.P. MORGAN ALTERNATIVE LOAN TRUST 2006-A1', we will add 'J.P. MORGAN ALTERNATIVE LOAN TRUST 2006-A1', 'J.P. MORGAN ALTERNATIVE LOAN', and 'MORGAN ALTERNATIVE LOAN' to the root dictionary, and 'TRUST 2006-A1' to the suffix dictionary.
\item
For a name that contains special tokens such as 'BANK', 'FUND', 'TRUST', etc.
we split the name into two parts.
We add the first fragment to the root dictionary and the last fragment to the
suffix directory.
However, if one of the fragments contains 'OF', add the fragment that includes the part with
the 'OF' to the root dictionary.
For example, from 'SAVINGS BANK OF THE FINGER LAKES FSB',
we will add 'OF THE FINGER LAKES FSB' to the root dictionary.
\item
For a long name with several tokens, compute tri-grams and add them to the root dictionary.
The tri-gram should not contain stop words such as 'THE', 'OF', etc.
\item
We utilize filters to remove tokens from the root and suffix dictionary.
For example, we use an address filter to remove 'STREET', 'CENTER', etc. and a
location filter to remove city names, etc.
\item
In addition to the tokens, our suffix dictionary contains regular expressions
that will mix and match tokens and numeric values.
For example, a financial institution that is set up as a special purpose vehicle
will have a name that includes the following:
'TRUST 2006-1', 'SERIES 2005-HE3', etc.
\end{itemize}

The above heuristics generated 354514 entries in the root dictionary and 26412
entries in the suffix dictionary.

\subsection{Dictionary Matching}
\label{sec:dict_mat}

After generating the root and suffix dictionary, we use the dictionary matching function
of System T ~\cite{chiticariu2010acl} to match the root and suffix of the extracted FI mentions.
The dictionary matching function is based on tokens.
It is robust and can handle line break characters
and other unexpected format issues.
We consider the root fragment in a mention as the unique and important fragment
and focus on a good match with entries in the root dictionary.
The suffix fragment is iteratively matched (and the suffix keyword is appended),
until we exhaust any possible additions from the suffix directory.

For further tuning, we include an additional customized root and suffix dictionary
and a dictionary of invalid elements.
These three dictionaries are customization points that can be used beyond
this specialized task to extract FI names.
The customized root and suffix dictionary are used in a similar
manner as the root and suffix dictionary that was automatically generated
from the list of FI names; this customization can further improve recall.
The dictionary of invalid elements is used to delete unexpected or potentially
incorrect mentions and can improve precision.

Despite the robust and effective dictionary matching function of System T,
we encountered the following difficult cases which require further effort.
\begin{itemize}
\item
Entity names that were split across several lines of text or across multiple columns.
This case is partially solved since the dictionary matching function
is based on tokens, and is insensitive to line breaks.
An example of an unsolved case is 'Wells \slash n abc def xxx \slash n Fargo Bank'.
\item
Entity names that were split into multiple fragments and where there were unrelated
sentences between the fragments of the entity name.
This case remains unsolved.
\end{itemize}

\subsection{Comparison of Dict-based NER and ORG NER}
\label{sec:compDictOrg}

\zheng{A reviewer suggest to move this to experiment, like what I did in the first draft. You could justify where is a proper place.}
ORG NER is a sophisticated general purpose rule-based NER tool that is also
built on the System T platform.
It achieved state-of-the-art performance on several standard NER tasks ~\cite{chiticariu2010emnlp}.
ORG NER has multiple customization points which are exposed as user-defined dictionaries.
These dictionaries allow ORG NER to be tuned for a variety of specialized domains. ORG NER encountered
several challenges when extracting mentions of FI names.
A majority of the errors involve the incomplete extraction of FI mentions;
the reasons are as follows:
\begin{itemize}
\item
First, the complex suffix templates for participant FI names in this dataset cannot be
easily captured by a suffix dictionary.  An example is a template that may contain a
date indicating when the prospectus was filed, or that may contain
a serial number for identification of the prospectus within a series.
An example is the following FI mention: "AAMES MORTGAGE INVESTMENT TRUST 20XX-Y",
where X and Y can be any digit value. For this scenario,
a regular expression based customization point for suffix identification may be appropriate.
\item
Next, even if an exhaustive dictionary of all suffix variations were available,
similar to the dictionary created by dict-based NER,
a general purpose NER may only achieve an incomplete match for the suffix.
ORG NER uses a complex set of rules to recognize and process a suffix.
The number of tokens, capitalization, and punctuation elements contained in a suffix
for FI mention would normally indicate the occurrence of
multiple named entity mentions (within the mention span)
or it may indicate the superset of a mention.
Handling such cases can confuse ORG NER rules and may prevent
ORG NER from completely utilizing the suffix dictionary entries.
This typically leads to an incomplete exraction
of the complete and complex suffix.
\item
Finally,
ORG NER relies on a complex combination of sentence boundary clues including whitespace,
newlines, punctuation, and capitalization to identify sentences.
It also makes the assumption that a named entity mention does not span multiple sentences.
Such clues for sentence identification and the heuristic for extracting a single mention
perform reasonably well for most unstructured text.
Unfortunately, both the sentence identification and the heuristic fail
when processing the header and summary sections of the resMBS prospecti for FI mentions.
This scenario, together with the two previous scenarios,
typically results in an incomplete extraction by ORG NER.
In particular, it will lead to an incomplete extraction of the suffix fragment of the FI name.
\end{itemize}

We summarize the comparison of ORG NER and Dict-based NER in Table ~\ref{tab:ner_comp}.

\zheng{Add an illustrative example?}

\begin{table}[tbhp]
\tbl{Comparison of Dict-based NER and ORG NER~\cite{chiticariu2010emnlp}, which are rule-based methods. We discuss the two methods in core rules, customized dictionaries and advantages, respectively.
\label{tab:ner_comp}}{
\begin{tabular}{c||p{5cm}|p{5cm}}
\hline
Comparison & Dict-based NER & ORG NER \\
\hline
\hline
Rules & After generation of root and suffix dictionary, straightforward matching and combination rule for extraction. Clear and transparent to users.   & Complex rules which consider keywords, context and format. Rules overlap with each other, which needs manual efforts to modify.  \\
\hline
Customization & Three customized dictionaries are exposed to users, which are used by matching rule. & Four customized dictionaries are exposed to users. Those dictionaries are used by sophisticated rules. \\
\hline
Advantages & Designed for financial institutions. Clear and transparent. Robust to document format. & Designed for general purpose. Rules are sophisticated, which use more information such as context. \\
\hline
\end{tabular}}%
\end{table}%

\section{Rank-based ER}
\label{sec:er}

The goal of the ER task is to resolve mentions and map (one or more) mentions
to a single financial institution (FI) name.
For example, "Wells Fargo Bank", "Wells Fargo Bank, N.A.", and "Wells Fargo Bank National Association"
should all (potentially) be mapped to "Wells Fargo".
Our specialized solution, Rank-based ER, exploits a corpus of names of FIs.
We assume that there exists a pre-defined corpus that has been normalized,
is targeted to this specialized task, and can cover a majority of mentions in the resMBS prospecti.
We use a corpus that was obtained from ABSNet\footnote{\url{http://www.absnet.net/ABSNet/}}.

We consider each FI name in this list to be a document, and we use a bag-
of-words model on the corpus of FI names for this task.
We develop a scoring function that is inspired by term frequency and inverse
document frequency (TF-IDF).
Rank-based ER also uses several heuristics based on the observed properties of FI names.
For each FI mention, the scoring function will be used to create a ranking and to find the best match from the corpus.
We use a threshold on the score to retain valid matches.
Rank-based ER uses an inverted index and is efficient and easy to parallelize.
Unlike NER, the root and suffix fragments of the name are not separated and are
incorporated into the scoring function.

\subsection{Index Construction}

The bag-of-words model uses an inverted index over the corpus of FI names.
To improve the efficiency of index search, we perform the following pre-processing steps
over both the query, i.e., the FI mention in the document, and the FI names in the corpus.
\begin{itemize}
\item We maintain a list of stop words; this includes words such as 'the' and more
specialized words such as 'LLC'.
We remove stop words and punctuation characters from the mentions.
\item  We maintain a mapping from abbreviations in the mentions to words or fragments in
the corpus.
For example, we map from 'WaMu' to 'Washington Mutual'.
\end{itemize}

\subsection{Scoring Function}
\label{sec:er_scoring}

A query corresponds to a mention of an FI in the document and is represented by
$\q=q_0 q_1 \ldots q_n$, where each $q_i$ is a token.  We create a candidate list
from the corpus of all FI names that include at least one $q_i$ and rank the list.
We use the following heuristics to develop the scoring function for ranking:
\begin{itemize}
\item Recall that an FI name comprises a root that is unique and a suffix.
The order of tokens in the FI name is important with the first few tokens being the
most important.
\item
If a candidate FI name from the corpus is a substring of the query, then there is a
high probability of a successful match from the query to the candidate.
\end{itemize}

Let $\p=\{ p_0 p_1 \ldots p_m \}$ represent a candidate name from the corpus, where $p_j$ is a word token.
We define a mapping function from $q$ to $p$ as follows:
\begin{eqnarray}
\text{map}(q_i, \p) =
\begin{cases} j, & q_i \in \q, p_j \in \p, q_i = p_j, \forall p_k = q_i, j \leq k,  \\
-1, & q_i \in \q, \forall p_k \in \p, q_i \neq p_k,  \\
\end{cases}
\end{eqnarray}

We identify the index $j$ or the j-th token of the candidate $\p$ that forms the first match for
token $q_i$ from the query.

We define an indicator function to signal if the query token $q_i$ exists in the candidate $\p$
as follows:
\begin{eqnarray}
\text{sgn}(q_i, \p) =
\begin{cases}
0, &\text{map}(q_i, \p) = -1,\\
1, &\text{map}(q_i, \p) \geq 0,\\
\end{cases}
\end{eqnarray}

We define a weight for each query token, $w(q_i)$, as the inverse document frequency (IDF) value.  This corresponds to the heuristic that the root fragment is typically unique and
is very important to a successful match.
We utilize a weight decay function $0.5^i$ to reflect the importance of the
(order of the ) first few tokens.
We also maintain a set of tokens for which we manually adjust the weight,
e.g, we reduce the weight for the token 'Structured' since it occurs in a moderate frequency, but is relatively non-informative.

We define a scoring function that consists of the following three factors:
(1) The first factor $s_{q}(\q, \p)$ corresponds to the weighted summation of all the matching tokens in the query $q$.
(2) The second factor $s_{c}(\q, \p)$ corresponds to the count of matching tokens in the query.
(3) the third factor $s_{b}(\q, \p)$ is a bonus when the candidate from the corpus is a substring of the query.

\begin{eqnarray}
s_{q}(\q, \p) 
= \sum_{i=0}^{i_{max}} 0.5^i*\text{sgn}(q_i, \p)*w(q_i)~\label{eq:sq}
\end{eqnarray}
where $i_{max} = \max_{i} \{i: \forall k<i, \text{map}(q_i, \p) > \text{map}(q_k, \p)\}$.

\begin{eqnarray}
s_{c}(\q, \p) 
= \frac{\sum_{j=0}^{j_{max}} 0.5^j*\text{sgn}(p_j, \q)}{\sum_{j=0}^{m} 0.5^j}~\label{eq:sc}
\end{eqnarray}
where $j_{max} =  \max_{j} \{j: \forall k<j, \text{map}(p_j, \p) > \text{map}(p_k, \p)\}$.

\begin{eqnarray}
s_{b}(\q, \p) 
=
\begin{cases}
\frac{\sum_{i=i_{min}}^{i_{min}+m} 0.5^i}{\sum_{i=0}^{n} 0.5^i}, &\exists i_{min}, \forall k \in \{0,1,\ldots m\}, q_{i_{min}+k} = p_k\\
0 & \text{otherwise}
\end{cases}.
\end{eqnarray}

The final scoring function combines the three factors as follows:
\begin{eqnarray}
\text{score}(\q, \p) = s_{q}(\q, \p)*s_{c}(\q , \p)+s_{b}(\q, \p)
\end{eqnarray}

We use a threshold on value of the scoring function to decide whether the mapped result
is valid. We determined a threshold through experiments and tuning and found threshold
of 0.085 worked well for the resMBS dataset.

\zheng{illustrative example for scoring function?}

\section{Experiments}
\label{sec:exp}

We evaluate the effectiveness of Dict-based NER and Rank-based ER to extract
names of financial institutions from a collection of over 5000 resMBS prospecti
that were filed with the SEC between 2000 and 2008.
Each document is uniquely labelled by the filing financial institution and
a unique identifier, the Central Index Key (CIK).
We note that there is no labeled training data available a priori, nor are there
multiple pre-populated dictionaries that could be customized.
Hence, all the dictionaries had to be constructed {\em from scratch} and we performed
an exhaustive manual evaluation, albeit with a limited number of documents.

We use Dict-based NER to recognize and extract the mentions of financial institutions and we
use Rank-based ER to link those extracted mentions
to a corpus that was obtained from ABSNet.
We used the general purpose ORG NER as a control for comparison with Dict-based NER.
We discuss the performance of Dict-based NER and ORG NER in Section ~\ref{sec:exp_ner},
and that of Rank-based ER in Section ~\ref{sec:exp_er}, respectively.

\subsection{Dict-based NER and ORG NER}
\label{sec:exp_ner}

We use a small number of randomly sampled documents for the evaluation.
We use 15 documents for dictionary construction as discussed in Section ~\ref{sec:dict_gen}.
We further use an additional 13 documents
\footnote{Those documents are from 9 institutions, AAMES, ABN\_AMRO, ABSC, ACE, AMERICAN\_HOME, BANC\_OF\_AMERICA, EAR\_STERNS, COUNTRYWIDE, and INDYMAC.}
to tune the customized dictionary described in Section ~\ref{sec:dict_mat}.
We use the same 28 documents to tune ORG NER.

For dictionary and index construction in Section \ref{sec:dict_gen},
we use multiple external sources, as discussed.
The SEC file contains 174851 names; however, only a small number of these
are FI names.
We also use a set of 6874 names from ORG NER.
Finally, a customized collection of about 50 FI names were extracted
from the 15 tuning documents.
Overall, we generate 354514 entries in the root dictionary and
26412 entries in the suffix dictionary.

The FI names are typically located in the header and summary sections of the resMBS prospectus.
Further, a financial institution that plays the role of an
{\em issuer} files the prospectus and may have a significant
impact on the selection of other FIs.  In order to perform an unbiased evaluation
we consider the following options:
\begin{itemize}
\item
We perform an evaluation of FI name mentions from the header and summary
sections, and also across the entire document.
We consider a collection of twenty three unseen test prospecti.
Of these 23 prospecti, we evaluate the extraction of mentions of FI names from the header and the
summary section for 18 documents and from the entire document for 5 documents.
\item
For the header and summary evaluation, the 18 unseen prospecti are from 12 institutions
\footnote{The twelve FI names are ABN\_AMRO, ACCREDITED, AMERICAN\_GENERAL, AMERICAN\_HOME, BANC
\_OF\_AMERICA, BEAR\_STERNS, COUNTRYWIDE, INDYMAC, LEHMAN, WACHOVIA, WAMU and WELLS\_FARGO.}.
Among these 12 institutions,  we did not consider training prospecti from four institutions
\footnote{The four FI names are ACCREDITED, AMERICAN\_GENERAL, LEHMAN and WACHOVIA.}
and this corresponded to five documents.
\item
We evaluate mentions from the entire document for 5 documents.
These five prospecti are filed (sponsored) by five institutions
\footnote{The five FI names are ABN\_AMRO, BANK\_OF\_AMERICA, BEAR\_STERNS, LEHMAN and WACHOVIA.}.
Among these five institutions and five documents,
we did not utilize prospecti from two institutions
\footnote{The two FI names are LEHMAN and WACHOVIA.}
as training prospecti; this corresponded to two documents from unseen FIs.
\end{itemize}

We consider the following measures:
\begin{itemize}
\item
{\bf ALL}:
This is the count of all the FI mentions that are extracted from the header and
summary sections.
\item
{\bf WRO}:
This refers to FI names that are extracted and then found to be completely incorrect,
e.g., the string {\tt May Be Limited By Book-Entry}.
\item
{\bf PAR}:
A partial extraction refers to FI mentions that have an overlap of tokens with the
correctly matching FI name.
For example, {\tt ABN AMRO} and {\tt Servicer ABN AMRO Mortgage} have an overlap
with the correct match {\tt ABN AMRO Mortgage Group, Inc.}.
The partial extraction results have a high probability to be correctly mapped to
the correct FI name in the corpus in the next step of Rank-based ER.
\item
{\bf MIS}:
A missing extraction refers to an FI mention in the document that is completely overlooked,
e.g., {\tt Second Street Funding}.
\end{itemize}

We consider all four measures for the task of FI name extraction from the
summary and header, but we only consider ALL and WRO for the human validation
task when we consider
the extraction of FI names from the entire text of the prospecti.
This is because each prospectus can typically include hundreds of pages of text;
reviewing all partial and missing extractions would take significant effort.
Despite these practical limitations, we believe that our evaluation and human
validation results are fairly robust and representative, as will be discussed.

We compute precision (PRE), recall (REC), partial precision (PAR\_PRE) and
partial recall (PAR\_REC).
The results labeled PAR are considered to be incorrect when computing
precision (PRE) and recall (REC).
They are considered to be correct when computing
partial precision (PAR\_PRE) and partial recall (PAR\_REc).
The values for precision and recall are calculated as follows:
\begin{eqnarray}
\text{PRE} &=& \frac{\text{ALL}-\text{WRO}-\text{PAR}}{\text{ALL}} \\
\text{PAR\_PRE} &=& \frac{\text{ALL}-\text{WRO}}{\text{ALL}} \\
\text{REC} &=& \frac{\text{ALL}-\text{WRO}-\text{PAR}}{\text{ALL}-\text{WRO}-\text{PAR}+\text{MIS}} \\
\text{PAR\_REC} &=& \frac{\text{ALL}-\text{WRO}}{\text{ALL}-\text{WRO}+\text{MIS}}
\end{eqnarray}
We also calculated the F1 score as follows:
\begin{eqnarray}
\text{F}_1 &=& 2*\frac{\text{PRE}*\text{REC}}{\text{PRE}+\text{REC}} \\
\text{PAR\_F}_1 &=& 2*\frac{\text{PAR\_PRE}*\text{PAR\_REC}}{\text{PAR\_PRE}+\text{PAR\_REC}}
\end{eqnarray}

\begin{table}[tbhp]
\tbl{All measures for Dict-based NER and ORG NER~\cite{chiticariu2010emnlp} for the extraction of FI names from the header and summary section for the 18 testing documents.
\label{tab:exp_ner_hs}}
{
\begin{tabular}{|c|c||c|c|c|c||c|c|c|c||c|c|}
\hline
Method & Section & ALL & WRO & PAR & MIS & PRE & PAR\_PRE & REC & PAR\_REC & F1 & PAR\_F1\\
\hline
\hline
Dict- & Header & 214 & 0 & 15 & 7 & 92.99\% & 100\% & 96.60\% & 96.83\% & .9476 & .9839\\
\cline{2-12}
based & Summary & 196 & 0 & 9 & 9 & 95.41\% & 100\% & 95.41\% & 95.61\% & .9541 & .9776\\
\cline{2-12}
NER & Both & 410 & 0 & 24 & 16 & 94.15\% & 100\% & 96.02\% & 96.24\% & .9508 & .9808\\
\hline
\hline
ORG & Header & 219 & 19 & 54 & 35 & 66.67\% & 91.32\% & 80.66\% & 85.11\% & .7300 & .8811\\
\cline{2-12}
NER & Summary & 208 & 9 & 53 & 14 & 70.19\% & 95.67\% & 91.25\% & 93.43\% & .7935 & .9454\\
\cline{2-12}
& Both & 427 & 28 & 107 & 49 & 68.38\% & 93.44\% & 85.63\% & 89.06\% & .7604 & .9120\\
\hline
\end{tabular}}%
\end{table}%

\begin{table}[tbhp]
\tbl{All measures for Dict-based NER and ORG NER~\cite{chiticariu2010emnlp}
for the extraction of FI names from the header and summary section for 5 out of
18 testing documents; these documents were sponsored (filed) by FIs where we did not
use prospecti filed by these FIs as training prospecti, i.e., unseen FIs.
\label{tab:exp_ner_uns}}
{
\begin{tabular}{|c|c||c|c|c|c||c|c|c|c||c|c|}
\hline
Method & Section & ALL & WRO & PAR & MIS & PRE & PAR\_PRE & REC & PAR\_REC & F1 & PAR\_F1\\
\hline
\hline
Dict- & Header & 57 & 0 & 6 & 7 & 89.47\% & 100\% & 87.93\% & 89.06\% & .8869 & .9421\\
\cline{2-12}
based & Summary & 46 & 0 & 0 & 9 & 100\% & 100\% & 83.64\% & 83.64\% & .9109 & .9109\\
\cline{2-12}
NER & Both & 103 & 0 & 6 & 16 & 94.17\% & 100\% & 85.84\% & 86.55\% & .8981 & .9279\\
\hline
\hline
ORG & Header & 66 & 1 & 16 & 6 & 74.24\% & 98.48\% & 89.09\% & 91.55\% & .8099 & .9489\\
\cline{2-12}
NER & Summary & 57 & 6 & 14 & 3 & 64.91\% & 89.47\% & 92.50\% & 94.44\% & .7629 & .9189\\
\cline{2-12}
& Both & 123 & 7 & 30 & 9 & 69.92\% & 94.31\% & 90.53\% & 92.80\% & .7890 & .9355\\
\hline
\end{tabular}}%
\end{table}%

\begin{table}[tbhp]
\tbl{
Partical precision and recall measures for
Dict-based NER and ORG NER~\cite{chiticariu2010emnlp}
for the extraction of FI names from the entire document.
We report on the results for all 5 documents/FIs (ALL FIs) and for the
2 out of 5 documents of unseen FIs (UNS FIs).
\label{tab:exp_ner_all}}
{
\begin{tabular}{|c|c||c|c||c|}
\hline
Method & Documents & ALL & WRO & PAR\_PRE\\
\hline
\hline
Dict-based & ALL FIs & 2912 & 4 & 99.86\% \\
\cline{2-5}
NER & UNS FIs & 1779 & 2 &  99.89\% \\
\hline
\hline
ORG & ALL FIs & 5208 & 623 & 88.04\% \\
\cline{2-5}
NER & UNS FIs & 2849 & 305 & 89.29\% \\
\hline
\end{tabular}}%
\end{table}%

We present the results of a human validation of the extraction of FI names
in Table ~\ref{tab:exp_ner_hs}, Table ~\ref{tab:exp_ner_uns} and Table ~\ref{tab:exp_ner_all}.
Table ~\ref{tab:exp_ner_hs} shows the results for Dict-based NER and ORG NER
for 18 testing documents.
We present results for the document sections, header and summary, both separately and together.
Table ~\ref{tab:exp_ner_uns} shows the results for 5 documents from unseen FIs,
i.e., we did not use training prospecti from these unseen FIs.
We observe that Dict-based NER demonstrates promising results.

The recall of Dict-based NER is comparable to that of ORG NER, while the
precision of Dict-based NER is consistently better.
To explain, ORG NER often misses the {\em issuing entity} from the prospectus.
The format for the FI name is typically {\tt XXX XXX Trust, Series XXXX-XXX}.
It often is a newly formed institution and appears in a single line in the header section.
ORG NER will also miss institutions, or extract a lot of partial results,
when several FI mentions appear in close vicinity of each other, separated by a comma.
An example is the list of mentions of FI names of resMBS servicers
as seen in Figure ~\ref{fig:intro}(a).

The precision of Dict-based NER is good for many reasons, e.g.,
dictionary construction discards some common tokens that may cause errors during
mention extraction.
Further, the dictionary matching step of Dict-based NER is robust to line breaks.
In contrast, ORG NER uses line breaks as a heuristic for extraction.
This decision by Dict-based NER significantly reduces the number of partial extractions.
In addition, ORG NER often uses heuristics, e.g., the use of capitalization,
which is not suitable for FI name extraction from resMBS prospecti.

We observe that Dict-based NER is robust and shows similar performance for both the
header and summary sections.  In contrast, ORG NER has greater variance across
the two sections and performs worse for the header; see Table ~\ref{tab:exp_ner_hs}.
To explain, the header section is more challenging since it is less well structured
and stylized and contains less contextual text that can be used for the extraction
of FI names.  In addition, contextual text may be misinterpreted and may lead to
incorrect FI name extraction.
We observe that this often happens for ORG NER with test documents
from unseen FIs, i.e., we did not include training prospecti from those FIs.

The recall of Dict-based NER drops when extracting FI mentions from entire prospecti. 
One challenge is the use of abbreviations of FI name mentions,
e.g., {\tt "WMC"} can represent {\tt "Wachovia Mortgage Corp."}.
Such abbreviations are localized to specific contracts filed by the related FI
and these mentions cannot be processed without some contextual text from the contracts.
We observe that in many cases, the abbreviation of the FI name will first be introduced
together with the full FI name in Summary and Header sections.  Subsequently, the abbreviations will be used
without the full FI name.
ORG NER has rules to handle many general cases and it has the capability
to cope with this challenge.
This is reflected in the correct number of extractions reported in Table ~\ref{tab:exp_ner_all}.

Dict-based NER generally performs well for both precision and recall.
As expected, it exhibits the best performance for FI name extraction when
processing test prospecti where the sponsoring (filing) institution has
previously provided training prospecti.
This can be used to benefit Dict-based NER since the most popular
(Top 15) sponsoring FIs file more than 80\% of the prospecti.
Thus, there are opportunities to further tune the performance of Dict-based NER.

To further understand the performance of Dict-based NER and ORG NER,
we consider their performance on partial precision ($\text{PAR\_PRE}$) and observe
that they show similar performance.
Both approaches had difficulty extracting complete FI name mentions.
The reasons for these partial FI name extractions were described in detail
in Section ~\ref{sec:compDictOrg}.

Finally, the more robust performance of the specialized domain specific
Dict-based NER, in comparison to the general purpose ORG NER, can be explained by
the ability to more easily tune Dict-based NER.
Dict-based NER is less complex and has fewer rules. It is able to benefit from
customizing the various dictionaries using the training prospecti.
This is reflected in the improved performance for the previously seen FIs
versus the unseen FIs.

To conclude, the experiment with documents from unseen institutions demonstrates
the generalizability of both approaches.
Further, when facing a scenario where less sample documents are available,
ORG NER showed higher recall but lower precision.
It also had almost identical F1 scores when considering partial precision and recall.

\subsection{Rank-based ER}
\label{sec:exp_er}

The evaluation of Rank-based ER is performed as follows:
\begin{itemize}
\item
Extract mentions of FI names from the header and summary sections of all 5131 prospecti.
Filter mentions to include those mentions that are adjacent to a keyword that may
indicate that the financial institution plays a specific role in the financial contract following~\cite{burdick2016dsmm}.
Example keywords from Figure \ref{fig:intro} are "Servicers", "Issuer" and "Sponsor".
This step yields 53354 mentions.
\item
Perform pre-processing and de-duplication to produce 5535 unique mentions of FI names.
\item
Find the best match for each mention against the ABSNet corpus of FI names;
there are 393 normalized names in this corpus.
Produce a tuple (unique FI name mention, ABSNet name, score) for each unique FI name mention.
\item
We rank the 5535 tuples by the mapping score, and draw the precision-recall curve as follows:
For some value of the threshold and for all tuples whose score is above this threshold,
we determine the count of true positives by manually checking the count
of correct mappings between the extracted FI names and the ABSNet corpus.
An extracted FI name that cannot be successfully mapped to the ABSNet corpus by Rank-based ER is considered
to be incorrect. 
The reasons for the incorrectness include both an incomplete ABSNet corpus as well as errors
during extraction of the FI name.
We use the count of true positives, i.e., the count of correct matching tuples
above the threshold, and the count of all tuples above the threshold,
to calculate precision and recall for a threshold. The precision-recall curve is generated by varying this threshold.
\item
We could empirically fix a feasible threshold by looking at the precision-recall curve. A threshold that achieves high precision and moderate recall is selected. After selecting the threshold, we evaluate on the 53354 non-deduplicated mentions.
In this case, we determine the count of true positives whose matching score
with the entry from the ABSNet corpus exceeds the threshold.  If the value is
lower than the threshold, then this is considered a mismatch.
\end{itemize}

\begin{figure}[tbhp]
\centerline{\includegraphics[width=\linewidth]{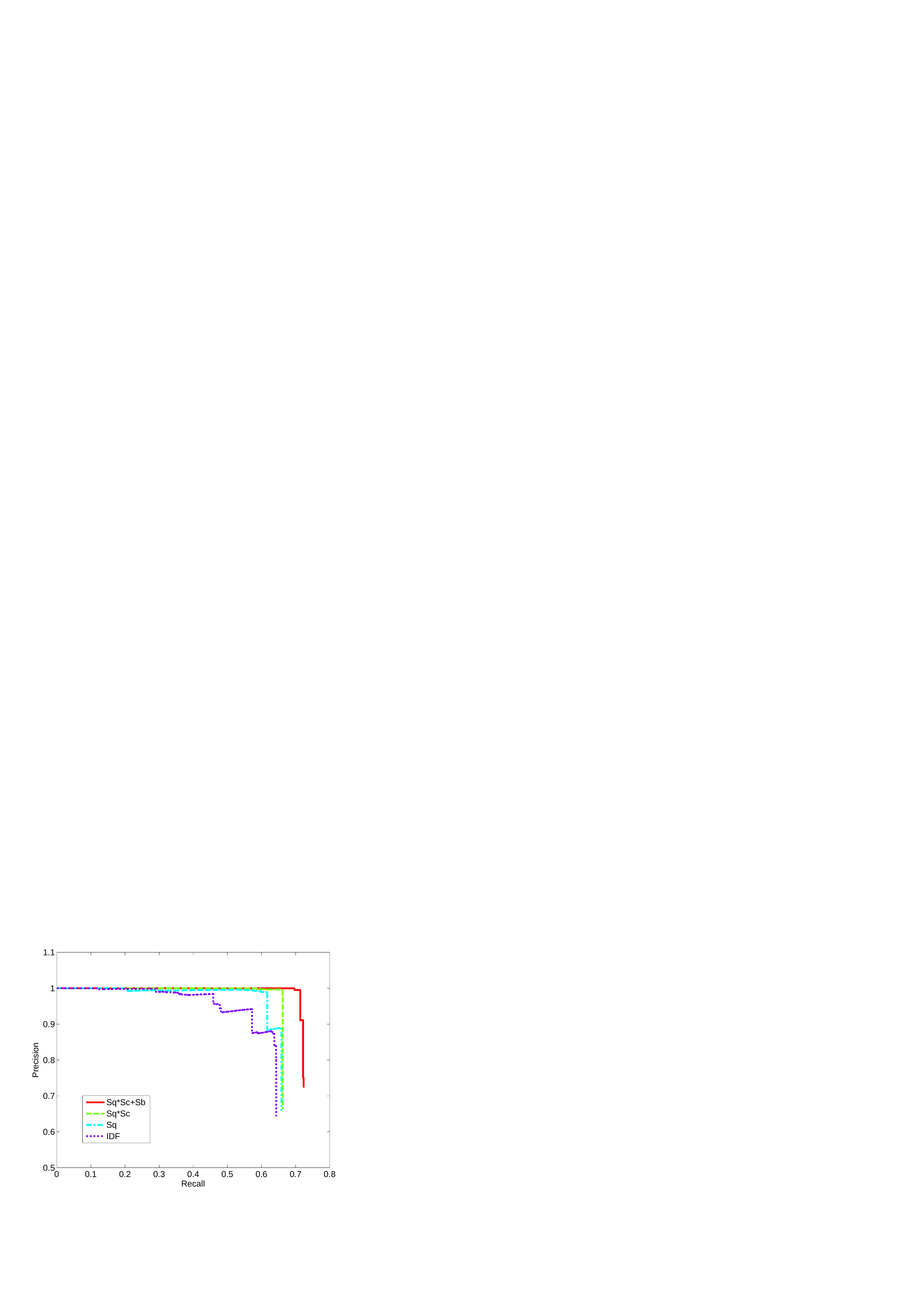}}
\caption{Precision-Recall Curve for Rank-based ER with Scoring Function $s_q*s_c+s_b$ and
Baseline Methods with Scoring Functions $s_q*s_c$, $s_q$ and IDF.}
\label{fig:pr}
\end{figure}

We compare the results of Rank-based ER against the following baseline methods:

\begin{itemize}
\item
A baseline that uses the IDF weight. Term frequency is usually 1 since the FI names are short, and hence is not considered.
\item
A baseline using the scoring function $s_q$ from Equation \ref{eq:sq};
it reflects the root and suffix heuristic of Section \ref{sec:er_scoring}.
\item
A baseline using the scoring function  $s_q*s_c$ from Equation \ref{eq:sc}
that also considers the order of the tokens discussed in Section \ref{sec:er_scoring}.
\end{itemize}

Figure \ref{fig:pr} shows the precision-recall curve for Rank-based ER;
it outperforms all three baselines.  The two baselines that consider
(simpler) scoring functions also outperform the IDF baseline.
\SAVE{
We compare the performance of the scoring function in rank-based ER with several baselines. We compare with scoring function using IDF weight, using only $s_q$ in Eq.~\ref{eq:sq} and using $s_q*s_c$ in Eq.~\ref{eq:sq} and Eq.~\ref{eq:sc}. The baseline methods are special cases of our proposed scoring function. The proposed method considers all the three heuristics in Sec.~\ref{sec:er_scoring}, while the baseline methods only considers some of the heuristics. Fig.~\ref{fig:pr} shows the precision-recall curve of rank-based ER. Our proposed scoring function outperforms the baseline methods. Baselines $s_q$ and $s_1*s_c$, which use both root-suffix heuristic (heuristic (1) in Sec.~\ref{sec:er_scoring}) and order heuristic (heuristic (2) in Sec.~\ref{sec:er_scoring}), performs better than the baseline IDF method. The results demonstrate the importance of heuristics in designing the scoring function and the efficiency of rank-based ER.
}

Consider the performance of Rank-based ER; it is labeled as $s_q*s_c+s_b$ in Figure \ref{fig:pr}.
Note that this figure reports on the results across the unique FI mentions.
The precision-recall curve shows that the precision maintains a consistently high value,
across a large range of threshold values.
We fixed the threshold at 0.085 for this set of experiments;
this resulted in a precision of 99.95\% and a recall of 69.66\%. We observe that almost all the issues in precision and recall are caused by either incomplete of ABSNet names or the extracted mentions are incorrect, which demonstrate the effectiveness of our scoring function.


Next, we consider the recall across all extracted FI mentions with the fixed threshold.
We report only on pseudo recall since we consider all the tuples with a score
above the threshold to be true positives instead of manually labelling 50000+ tuples. Since the precision of Rank-based ER is almost 100\%, the pseudo recall is representative. 
We obtain a much higher value for pseudo recall of 88.27\%.
To explain, when considering all mentions, the popular financial institutions that
participate in many resMBS contracts typically can find a match in the ABSNet corpus.
These popular FI names may appear multiple times across different contracts.
In addition, if there is an incorrect extraction, the error in the FI name
will only be recorded once and the error will not be duplicated.

To further improve the recall, we dig into details of the existing issues in rank-based ER.
We observe that several issuer institutions can not find a mapping in ABSNet, such as "HSI Asset Securitization Corporation Trust XXXX-XXXX", "MASTR ADJUSTABLE RATE MORTGAGES TRUST XXX-X",
"RALI SERIES XXXX-XXX TRUST ", "RASC SERIES XXXX-XXX TRUST", "Citicorp Mortgage Securities Trust Series XXXX-X",
and "RAMP SERIES XXXX-XXX TRUST". For those issuers, we can not map the name abbreviation to an informative institution name. We should extend our normalized list in rank-based ER to include those abbreviations. However, it is a nontrivial task since those names are not common names and are also difficult to recognize for experts.

 Moreover, some examples of extraction could not possibly recognized are, "ALTERNATIVE LOAN TRUST XXXX-XXXX", and "ASSET BACKED NOTES SERIES XXXX-X". Those names cannot be recognized without context information and cannot be solved by the rank-based ER framework. We need to go back to the documents to find relevant descriptions for those names to correctly recognize them.

%

\section{Conclusion}

We proposed a specialized rule-based solution for the extraction of FI names.
Our innovation is to exploit lists of FI names,
and to customize a two-part solution, Dict-based NER and Rank-based ER.
Dict-based NER and Rank-based ER are built upon the algebraic information
extraction system, System T.
Dict-based NER can be viewed as a specialization of the general purpose ORG NER that
is also available on the System T platform.

We combine multiple lists of FI names from several sources for Dict-based NER.
In contrast, we utilize a smaller targeted list of names for Rank-based ER.
We observe that FI names can typically be split into a root fragment and a suffix.
We generate a root dictionary and a suffix dictionary from the lists of FI names,
and Dict-based NER will utilize a dictionary matching function to perform extraction.
The root and suffix dictionaries can synergistically help both modules
in extracting the root and the suffix.
For Rank-based ER, we develop a scoring function to select the best
matches against a corpus of FI entity names.

We evaluate the effectiveness of Dict-based NER and Rank-based ER to extract
FI names from a collection of over 5000 resMBS prospecti and we compare
Dict-based NER with ORG NER.
The recall of Dict-based NER is comparable to that of ORG NER, while the
precision of Dict-based NER is consistently better.
To explain, there are several cases where ORG NER will miss FI names or will
extract partial results.
Dict-based NER is helped by the root and suffix dictionaries and other heuristics
to avoid these cases.
We observe that Dict-based NER is robust and shows similar performance for both the
header and summary sections.  In contrast, ORG NER has greater variance across
the two sections and performs worse for the header.
To explain, the header section is more challenging since it is less well structured
and stylized and contains less contextual text that can be used for the extraction
of FI names.  In addition, contextual text may be misinterpreted and may lead to
incorrect FI name extraction.

There are several lessons learned from this experience that can be used to
improve upon our current solutions and to develop solutions for other specialized
NER and ER tasks.

Our first lesson is that a general purpose NER such as ORG NER will benefit from
more extensive dictionaries to capture domain and task specific knowledge.
These could be created using approaches similar to those used for Dict-based NER.
An example of an external list of FI names would be the names of all financial
institutions that have been issued a CIK. Another example is the use of the
root and suffix dictionaries.

Our next lesson goes beyond the dictionary based customization discussed in the paper.
A general purpose NER such as ORG NER may benefit from additional types of customization points.
Recall that the names of issuers of the resMBS contracts, issuer FI names,
were FI names that had been further modified.
A potential solution would be to include a regular expression based customization that
would similarly extend FI names and recognize the names of issuer FIs.

Our final lesson is very positive since we believe that both the general purpose
ORG NER and the special purpose Dict-based NER and Rank-based ER
can be applied with additional minimal customization to a range of other collections.
This includes the prospecti for other classes of asset backed securities,
e.g., ABS that are created by pooling auto loans, student loans, etc.
Prospectus documents for asset backed securities share similar formatting templates
as the resMBS prospecti, e.g., relevant information is captured in a header or
summary section.
Additionally, FI names for financial entities participanting in ABS prospecti
follow similar naming conventions, e.g., using a root and suffix.
Applying our techniques for dict-based NER and rank-based ER to the wider class of
ABS prospectuses remains as future work.

\begin{acks}
The authors would like to thank Soham De, Howard Ho, Rajasekar Krishnamurthy and Michael Shao for their feedback.
\end{acks}

\bibliographystyle{ACM-Reference-Format-Journals}
\bibliography{NER+ER}

\received{February 2016}{}{}



\end{document}